\documentclass[10pt,twocolumn,letterpaper]{article}

\usepackage{cvpr}
\usepackage{times}
\usepackage{epsfig}
\usepackage{graphicx}
\usepackage{amsmath}
\usepackage{amssymb}
\usepackage{color}
\usepackage[export]{adjustbox}
\newcommand{\todo}[1]{}
\renewcommand{\todo}[1]{{\color{red} TODO: {#1}}}
\newcommand{\alekstodo}[1]{}
\renewcommand{\alekstodo}[1]{{\color{red} ALEKS TODO: {#1}}}
\newcommand{\aleks}[1]{}
\renewcommand{\aleks}[1]{{\color{blue}{#1}}}

\usepackage{gensymb}

\usepackage[pagebackref=true,breaklinks=true,letterpaper=true,colorlinks,bookmarks=false]{hyperref}

 \cvprfinalcopy 

\def\cvprPaperID{3829} 

\ifcvprfinal\fi
\begin{document}

\title{Structure from Motion for Panorama-Style Videos}

\author{Chris Sweeney\thanks{The majority of this work was performed while the author was at the University of Washington.}\\
Facebook Reality Labs\\
{\tt\small sweeneychris@fb.com}
\and
Aleksander Holynski
\qquad
Brian Curless 
\qquad
Steve M. Seitz  \\
University of Washington \\
{\tt\small \{aholynski, curless, seitz\}@cs.washington.edu}
}


\maketitle

\begin{abstract}
We present a novel Structure from Motion pipeline that is capable of reconstructing accurate camera poses for panorama-style video capture without prior camera intrinsic calibration. While panorama-style capture is common and convenient, previous reconstruction methods fail to obtain accurate reconstructions due to the rotation-dominant motion and small baseline between views. Our method is built on the assumption that the camera motion approximately corresponds to motion on a sphere, and we introduce three novel relative pose methods to estimate the fundamental matrix and camera distortion for spherical motion. These solvers are efficient and robust, and provide an excellent initialization for bundle adjustment. A soft prior on the camera poses is used to discourage large deviations from the spherical motion assumption when performing bundle adjustment, which allows cameras to remain properly constrained for optimization in the absence of well-triangulated 3D points. To validate the effectiveness of the proposed method we evaluate our approach on both synthetic and real-world data, and demonstrate that camera poses are accurate enough for multiview stereo.

\end{abstract}

\section{Introduction}
\begin{figure}[t]
\centering
\includegraphics[width=1.0\linewidth,keepaspectratio]{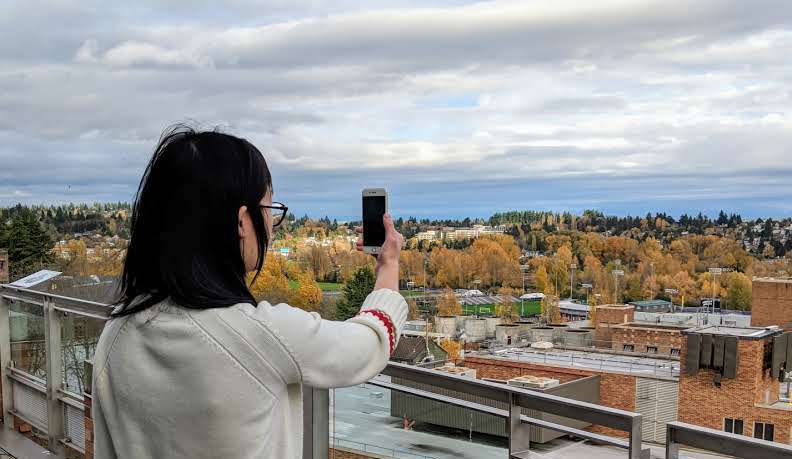}
\caption{\label{fig:pano_capture} Our system is able to successfully compute camera pose and geometry for panorama-style video sequences from handheld cameras. Typical SfM systems fail to reconstruct these scenes due to the rotation-dominant motion and limited baseline between views.}
\vspace{-0.1in}
\end{figure}

The way that we capture and share experiences is constantly evolving. With the rise of social media platforms and the recent increase in camera quality on mobile phones, we are no longer limited to only sharing images. Panoramas provide a convenient method for increasing the visual coverage of the scene and utilize the same visualization as standard images. Capturing a panorama is simple and convenient: you simply rotate the camera in your outstretched arm to scan the scene of interest (\cf Figure \ref{fig:pano_capture}). This type of capture is a natural motion, just as easy as taking a video, and captures far more visual information about the surrounding scene than an image alone.


While a panorama provides more visual information than a standard image, it still does not truly convey the feeling of ``being there" because it does not properly capture parallax. To achieve greater immersion, a 3D representation of the scene is necessary to provide stereo and depth cues as we would experience in the real world. Ideally, this 3D information could be incorporated into a panoramic 3D model of your surroundings allowing for depth cues when parallax can be detected and seamlessly reverting to a standard panorama in areas without parallax. In order to detect parallax, however, camera poses must be accurately reconstructed \eg, with Structure-from-Motion (SfM) or Simultaneous Localization and Mapping (SLAM) techniques.

Unfortunately, scenes that are well-suited for panorama capture are often ill-conditioned for reconstruction with conventional SfM or SLAM methods. This is because the rotation-dominant motion leads to limited overlap between images, thus the baseline between images that contain common features is typically small relative to the scene depth and point triangulation for reconstruction is unstable. For cell phone cameras, this problem is further amplified by the limited field of view of the camera and additional challenges arise due to motion blur and rolling shutter. Capturing image sequences with sufficient baseline is of course possible, but it is an imperfect and unnatural process especially for untrained users. As such, it is difficult to capture a scene with the proper constraints to obtain high quality SfM reconstructions without some form of online feedback.



In this paper, we present a system for reconstructing handheld videos that provides the convenience of panorama-style capture with the reconstruction quality of traditional SfM pipelines. Our accurate pose reconstructions are key to enabling realistic 3D scene reconstruction and visualization via multiview stereo and image-based rendering methods.
Our method does not require prior camera calibration, and is able to efficiently obtain accurate reconstructions despite the rotation-dominant motion of panorama-style capture. A core contribution of this work is three novel pose methods that model panorama-style motion as motion on a sphere~\cite{ventura2016structure} and allow us to obtain accurate camera intrinsic calibration and relative motion between video frames. We use these solvers to initialize an SfM pipeline that exploits priors on the camera motion during panorama-style capture. These priors allow our system to be robust to a variety of scenes, including scenes with distant scene points that are typically unsuitable for traditional reconstruction techniques. The system is fast, allowing captures in roughly a minute, and works on a number of scenes where state-of-the-art SfM and SLAM pipelines struggle or completely fail. We test our solvers and the SfM pipeline on several video sequences from a Google Pixel and GoPro, and demonstrate the applicability of our pipeline for enabling VR experiences.

\section{Related Work}
Inferring the relative geometry between two images is a core problems in geometric computer vision. Point correspondences between images are typically used to compute the pose, with the number of correspondences required often corresponding to the degrees of freedom of the problem. So called ``minimal solvers" have received much attention in recent years due to their applicability for pose estimation in the presence of noise and outliers in RANSAC schemes~\cite{bujnak2012making, byrod2009minimal, fischler1981random, kneip2011novel, kukelova2012polynomial}. Specialized algorithms have been created that exploit specific camera motions\cite{fitzgibbon1998automatic, hernandez2007silhouette, scaramuzza2009real, ventura2016structure}, or available sensor measurements (\eg internal measurement units) that reduce the degrees of freedom for solving for camera pose~\cite{sweeney2014solving, sweeney2015efficient}. Our work belongs in the former category as we exploit the fact that panorama-style motion can be approximately modeled as motion on a sphere to derive efficient minimal solvers.

Constraining motion to lie on a sphere was first exploited by Ventura~\cite{ventura2016structure} to determine the essential matrix (and thus relative pose) between two images with known camera calibration for this particular motion. The assumption of spherical motion reduces the degrees of freedom in the problem from 5 for the general case~\cite{nister2004efficient} to 3, enabling an efficient and accurate algorithm for determining relative motion. The author demonstrates the applicability of this solver in a spherical SfM system, though he notes that the accuracy for outward-facing sequences can be poor and deviations from the spherical motion assumption cause the system to fail. 

We build on Ventura's work~\cite{ventura2016structure} to design three new solvers for determining the fundamental matrix and calibration when the motion between cameras is constrained to a sphere. Whereas the fundamental matrix for general motion is estimated from 7~\cite{hartley2003multiple} or 8-point correspondences~\cite{hartley1997defense}, our spherical motions solvers compute the fundamental matrix from as few as 4 point correspondences. We can additionally compute a common radial distortion parameter with 5 or 6 correspondences. These fundamental matrix solvers are integrated into an SfM pipeline suitable for reconstructing panorama-style captures from handheld video sequences. Unlike ~\cite{ventura2016structure}, our method does not require prior calibration, allowing for more general use, and our system is robust to deviations from spherical motion. 





\section{Spherical Geometry for Handheld Video}
\label{sec:spherical_geometry}

\begin{figure}[t]
\centering
\includegraphics[width=0.4\linewidth,keepaspectratio]{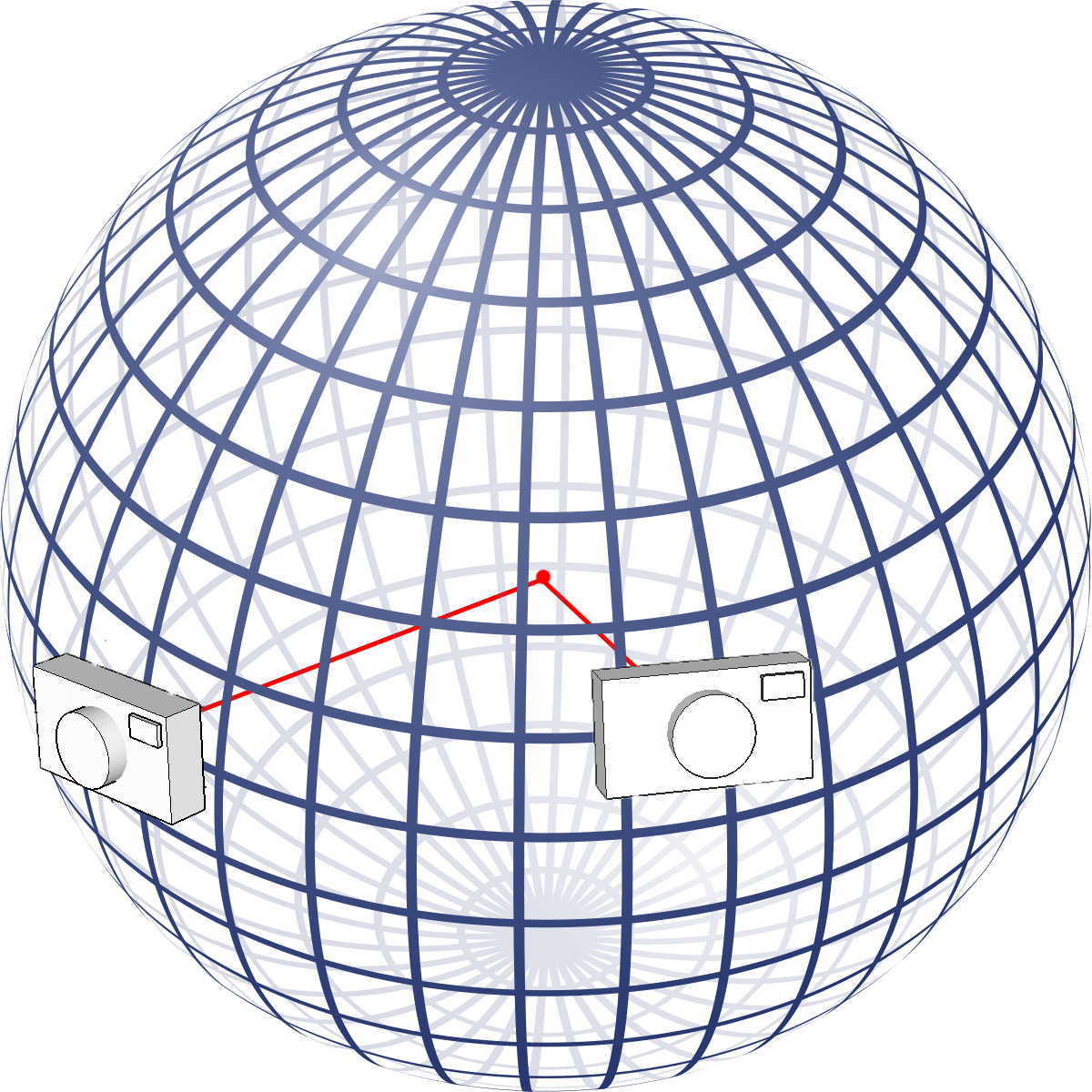}
\caption{\label{fig:spherical_capture} For spherical motion, cameras rotate at a fixed distance from the sphere's center with the optical axis aligned to the ray from the camera center to the sphere center. This reduces the degrees of freedom of the camera pose from 6 to 3, allowing for simplified methods to determine camera pose.}
\vspace{-0.1in}
\end{figure}

Our pano-style capture motion roughly follows a ``spherical motion'' trajectory~\cite{ventura2016structure} where the camera rotates at a fixed distance from an origin and the optical axis is aligned with the ray between the origin and camera center (\cf Figure \ref{fig:spherical_capture}). The camera faces outward towards the scene, with a possible additional rotation about its optical axis. Under spherical motion the camera position is exactly defined by the camera rotation, leading to only 3 degrees of freedom for the camera pose. The global scale of the scene is defined by the radius of the sphere and, further, the \textit{relative scale} between camera pairs is known exactly because the radius of the sphere is fixed. For an outward facing camera under spherical motion, the camera pose may be expressed as:

\begin{equation}
P = K\left[R|-z\right],
\end{equation}
where $K$ is the camera intrinsic matrix, $R$ is the rotation matrix and $z = \left[0 \; 0 \; 1\right]^\top$. The intrinsic matrix is defined as:

\begin{equation}
K =
\begin{bmatrix}
f & s & p_x \\
0 & \alpha f & p_y \\
0 & 0 & 1
\end{bmatrix},
\end{equation}
where $f$ is the focal length of the camera, $(p_x, p_y)$ is the principal point, $s$ is the skew and $\alpha$ is the pixel aspect ratio. We assume that the principal point is located at the center of the image, and that the skew is 0 with a unit pixel aspect ratio. This leads to a simplified intrinsic matrix:
\begin{equation}
K =
\begin{bmatrix}
f & 0 & 0 \\
0 & f & 0 \\
0 & 0 & 1
\end{bmatrix}.
\end{equation}

The assumption of spherical motion presents several significant advantages over general motion. First, the relative poses between camera pairs may be directly composed together in a common coordinate system because the scale of the relative motion is known. Second, the reduced degrees of freedom (3, as opposed to 5 for general motion) leads to simplified relative pose methods~\cite{ventura2016structure}. Third, the relative pose methods for spherical motion are robust to small motion and very distant scene depth due to the known scale.

\subsection{Epipolar Geometry for Spherical Motion}
The essential matrix relates normalized image correspondences $x_n$ and $x_n'$ (\ie, with the effect of camera intrinsic parameters removed) via the epipolar constraint:
\begin{equation}
x'^\top E x = 0.
\end{equation}
For motion on a sphere, Ventura~\cite{ventura2016structure} showed that the essential matrix has a special form due to the constrained motion:
\begin{equation}
\label{eq:essential_matrix}
E =\begin{bmatrix}
e_1 & e_2 & e_3 \\
e_2 & -e_1 & e_4 \\
e_5 & e_6 & 0
\end{bmatrix},
\end{equation}
The six unknown variables of the essential matrix may be efficiently determined from three normalized point correspondences~\cite{ventura2016structure}.

The essential matrix assumes explicit knowledge of the camera intrinsic calibration. This information may be unavailable (most cell phone cameras do not store focal length information in video metadata) or difficult to obtain. In the absence of camera intrinsic information, the fundamental matrix $F$ may instead be use to describe the epipolar geometry where the relation
\begin{equation}
E = K^\top F K,
\end{equation}
relates the essential matrix to the fundamental matrix with the camera intrinsics matrix K. We assume the the two images share the same camera intrinsics.

\begin{equation}
F = K^{-\top} E K^{-1} = \begin{bmatrix}
e_1 & e_2 & fe_3 \\
e_2 & -e_1 & fe_4 \\
fe_5 & fe_6 & 0
\end{bmatrix} = \begin{bmatrix}
f_1 & f_2 & f_3 \\
f_2 & -f_1 & f_4 \\
f_5 & f_6 & 0
\end{bmatrix},
\end{equation}
where the inverse of the camera intrinsics matrix is parameterized as
\begin{equation}
K^{-1} = K^{-\top} = \begin{bmatrix}
1 & 0 & 0 \\
0 & 1 & 0 \\
0 & 0 & f
\end{bmatrix}.
\end{equation}

Note that while 6 unknowns are used to represent the fundamental matrix, there are only 4 degrees of freedom (3 for rotation, 1 for focal length). As such, only 4 point correspondences are necessary to compute the fundamental matrix as opposed to 7 for the case of general motion~\cite{hartley2003multiple}.

\subsection{Estimating the Fundamental Matrix}
Given $N \geq 4$ point correspondences, the epipolar constraints for the fundamental matrix for spherical motion are given by:
\begin{equation}
\label{eq:fmatrix_epipolar_constraint}
p_i'^\top F p _i= 0, \;\;\; i=1,\ldots, N.
\end{equation}
Each point correspondence gives rise to a single linear equation in the entries of $F$. We can rearrange the epipolar constraint to obtain a linear system:
\begin{equation}
A \hat{F} = 0,
\end{equation}
where $A$ is a $N$ $\times$ 6 coefficient matrix and \mbox{$\hat{F} = \left[f_1 \; f_2 \; f_3 \; f_4 \; f_5 \; f_6\right]^\top$}. By computing the singular value decomposition of $A$, we may express $F$ in terms of two vectors spanning the right nullspace of $A$:
\begin{equation}
\label{eq:fundamental_one_scalar}
F = xF_1 + (1 - x) F_2,
\end{equation}
for some scalar $x$. To solve for $x$, we may substitute Eq.~\eqref{eq:fundamental_one_scalar} into the singularity constraint of the fundamental matrix:
\begin{equation}
\label{eq:singularity_constraint}
det(F) = 0.
\end{equation}
Similar to the seven point algorithm for computing the fundamental matrix for general motion~\cite{hartley2003multiple}, this leads to a cubic polynomial in $x$, whose roots reveal up to 3 solutions for the fundamental matrix.
 
\subsection{Incorporating Radial Distortion}
In practice, virtually all cameras exhibit some amount of radial distortion. Fitzgibbon~\cite{fitzgibbon2001simultaneous} demonstrated that ignoring radial distortion can lead to significant errors in camera calibration and feature matching, even for standard consumer cameras. Further, radial distortion can introduce many local minima to bundle adjustment problems, making a good initialization of camera parameters especially important~\cite{fitzgibbon2001simultaneous}. Properly modeling radial distortion is especially important for wide field-of-view cameras such as GoPros.

To model the effect of radial distortion we use the one-parameter radial \textit{undistortion} model presented in ~\cite{fitzgibbon2001simultaneous}. In this model, undistorted points are related to their distorted counterparts as:
\begin{equation}
\label{eq:distortion}
p_u(\lambda) = \left[x_d, y_d, 1 + \lambda(x_d^2 + y_d^2) \right]^\top,
\end{equation}
where $p_u$ is the homogeneous coordinate of the undistorted point, $p_d = \left(x_d, y_d \right)$ is the distorted point, and $\lambda$ is the radial distortion parameter. This representation of undistorted points may be susbtituted into the epipolar constraint of Eq.~\eqref{eq:fmatrix_epipolar_constraint}:
\begin{equation}
\label{eq:epipolar_and_distortion}
\begin{bmatrix} x_d' \\ y_d' \\ 1 + \lambda r_d'^2\end{bmatrix}^\top
\begin{bmatrix}
f_1 & f_2 & f_3 \\
f_2 & -f_1 & f_4 \\
f_5 & f_6 & 0
\end{bmatrix}
\begin{bmatrix} x_d \\ y_d \\ 1 + \lambda r_d^2 \end{bmatrix}
=0,
\end{equation}
where $r_d^2 = x_d^2 + y_d^2$ and $r_d'^2 = x_d'^2 + y_d'^2$. Again, we assume that the two images share the same camera intrinsics and thus have the same radial distortion. 

When accounting for radial distortion there are now 5 degrees of freedom for spherical motion (3 for rotation, 1 for focal length, and 1 for radial distortion), implying that we can solve for the fundamental matrix and radial distortion parameter from 5 point correspondences. Given 5 point correspondences, we can reform Eq.~\eqref{eq:epipolar_and_distortion} into a linear system:
\begin{equation}
\label{eq:linear_distortion}
AX = 0
\end{equation}
where $A$ is a 5 $\times$ 10 coefficient matrix with and \mbox{$X = \left[ \lambda f_3, \lambda f4, \lambda f_5, \lambda f6, f_1, f_2, f_3, f_4, f_5, f_6 \right]^\top $}

After applying Gaussian elimination to matrix $A$, one can linearly eliminate 5 unknown monomials by expressing them in terms of the remaining 5 monomials. Since $f_1$ and $f_2$ appear linearly in Eq.~\eqref{eq:linear_distortion}, it is natural to eliminate them. We additionally choose to eliminate $\{f_3, \lambda f_3, \lambda f_4\}$. The eliminated monomials $\{f_1, f_2, f_3, \lambda f_3, \lambda f_4 \}$ may now be expressed in terms of the remaining monomials $\{f_4, f_5, f_6, \lambda f_5, \lambda f_6\}$, or equivalently as a quadratic function of the unknown variables $\{f_4, f_5, f_6, \lambda\}$. Thus, for $f_1,f_2,f_3$ we have:
\begin{equation}
\label{eq:f_elimination}
f_i = h_i(f_4, f_5, f_6, \lambda)
\end{equation}
For monomials $\lambda f_3$ and $\lambda f_4$ we have
\begin{align}
\label{eq:eliminate_f3}
\lambda f_3 &= h_4(f_4, f_5, f_6, \lambda) \\
\label{eq:constraint1}
\lambda f_4 &= h_5(f_4, f_5, f_6, \lambda)
\end{align}
We can further eliminate $f_3$ from Eq.~\eqref{eq:eliminate_f3} by replacing it with $h_3(f_4, f_5, f_6, \lambda)$:
\begin{equation}
\label{eq:constraint2}
\lambda h_3(f_4, f_5, f_6, \lambda) - h_4(f_4, f_5, f_6, \lambda) = 0
\end{equation}
Additionally, we can substitute the expressions of ~\eqref{eq:f_elimination} into the singularity constraint of Eq.~\eqref{eq:singularity_constraint}:
\begin{equation}
det\left( \begin{bmatrix}
h_1 & h_2 & h_3 \\
h_2 & -h_1 & f_4 \\
f_5 & f_6 & 0
\end{bmatrix} \right) = 0
\end{equation}
Together with Eq.~\eqref{eq:constraint1} and Eq.~\eqref{eq:constraint2}, we obtain 3 equations in 4 unknowns. Since the fundamental matrix may only be estimated up to scale, we may obtain a 4th equation by constraining the norm of the fundamental matrix:
\begin{align}
||F||^2 - 1 &= 0 \\
2h_1^2 + 2h_2^2 + h_3^2 + f_4^2 + f_5^2 + f_6^2 - 1 &= 0
\end{align}
To avoid any square roots, we constrain the \textit{squared} norm of the fundamental matrix. 

We use the automatic tool from the Polyjam software library\footnote{\url{https://github.com/laurentkneip/polyjam}}~\cite{kneip2014opengv} to find a reduced Gr\"obner basis for solving the system of 4 polynomials described\footnote{For a detailed description of Gr\"obner basis techniques and their applications to computer vision, please see~\cite{byrod2007improving, kukelova2008automatic, stewenius2005grobner}} The solver performs row reduction via LU decomposition on a 211 $\times$ 233 coefficient matrix, producing a 20 $\times$ 20 action matrix from the coefficients of the reduced coefficient matrix. The eigenvectors of this action matrix reveal up to 20 solutions for $\{f_4, f_5, f_6, \lambda\}$, from which only the real solutions are kept. We have found that typically 8 or fewer real solutions exist in practice.

\subsection{An Efficient Generalized Eigenvalue Solution}
The Gr\"obner basis method presented in the previous section is inefficient due to the large LU decomposition and thus is expensive to use in a RANSAC scheme~\cite{fischler1981random}. To design a more efficient solution method, we follow the idea of Kukelova \etal ~\cite{kukelova2013real} and use an additional correspondence to obtain a dramatically simpler system of equations that can be solved more efficiently and, in our case, with greater numerical stability.

If we consider 6 point correspondences instead of the minimal 5, the linear system of Eq.~\eqref{eq:linear_distortion} formed from the epipolar constraint may instead be written as:
\begin{equation}
(\lambda C_1 + C_2) \bold{f} = 0,
\end{equation}
where $C_1$ and $C_2$ are 6 $\times$ 6 coefficient matrices formed directly from the epipolar constraint and \mbox{$f = \left[f_1 \; f_2 \; f_3 \; f_4 \; f_5 \; f_6 \right]^\top$}. This linear system is a Generalized Eigenvalue Problem (GEP) of the form:
\begin{equation}
C_2 f = -\lambda C_1 f,
\end{equation}
which may be readily solved with linear algebra software packages such as Eigen\footnote{\url{http://eigen.tuxfamily.org/}}. After solving this eigenvalue problem, the eigenvalues of this GEP correspond to the solutions for the radial distortion parameter $\lambda$, and the corresponding eigenvectors contain the entries of $F$. 

\section{Uncalibrated Reconstruction on a Sphere}
The methods described in Section~\ref{sec:spherical_geometry} allow us to solve for radial distortion parameters and fundamental matrices between two cameras from point correspondences. In this section, we describe a Structure from Motion pipeline which uses these minimal solvers to determine camera poses, calibration, and scene structure from uncalibrated video input. The input to our pipeline is an outward-facing video sequence undergoing spherical motion. Compared to the pipeline of Ventura~\cite{ventura2016structure}, our system does not require camera calibration and is more robust to deviations from spherical motion.

At a high level, our system tracks features through the input video and estimates fundamental matrices and radial distortion between successive keyframes of the video sequence. Once the entire sequence has been tracked, we determine camera calibration parameters (\ie, focal length) by using ``infinite depth" constraints from distant features in the scene. This calibration can then be applied to the fundamental matrices to obtain essential matrices, and thus relative pose parameters~\cite{ventura2016structure} between keyframes. A global rotation averaging method is used to determine camera orientations in a global coordinate system. The positions of camera undergoing spherical motion are exactly determined from the camera orientations, which provides a good initialization for bundle adjustment.

\subsection{Feature Tracking}
We first track feature points detected with OpenCV's GoodFeaturesToTrack~\cite{shi1994good} using Kanade-Lucas-Tomasi optical flow. Keyframes are created when the mean optical flow distance exceeds $0.02 *max(image\_width,\;image\_height)$ pixels, and the fundamental matrix (and optionally radial distortion) between keyframes is robustly estimated with MLESAC. 

\subsection{Calibration}
\begin{figure}[t]
\centering
\includegraphics[width=0.495\linewidth,keepaspectratio]{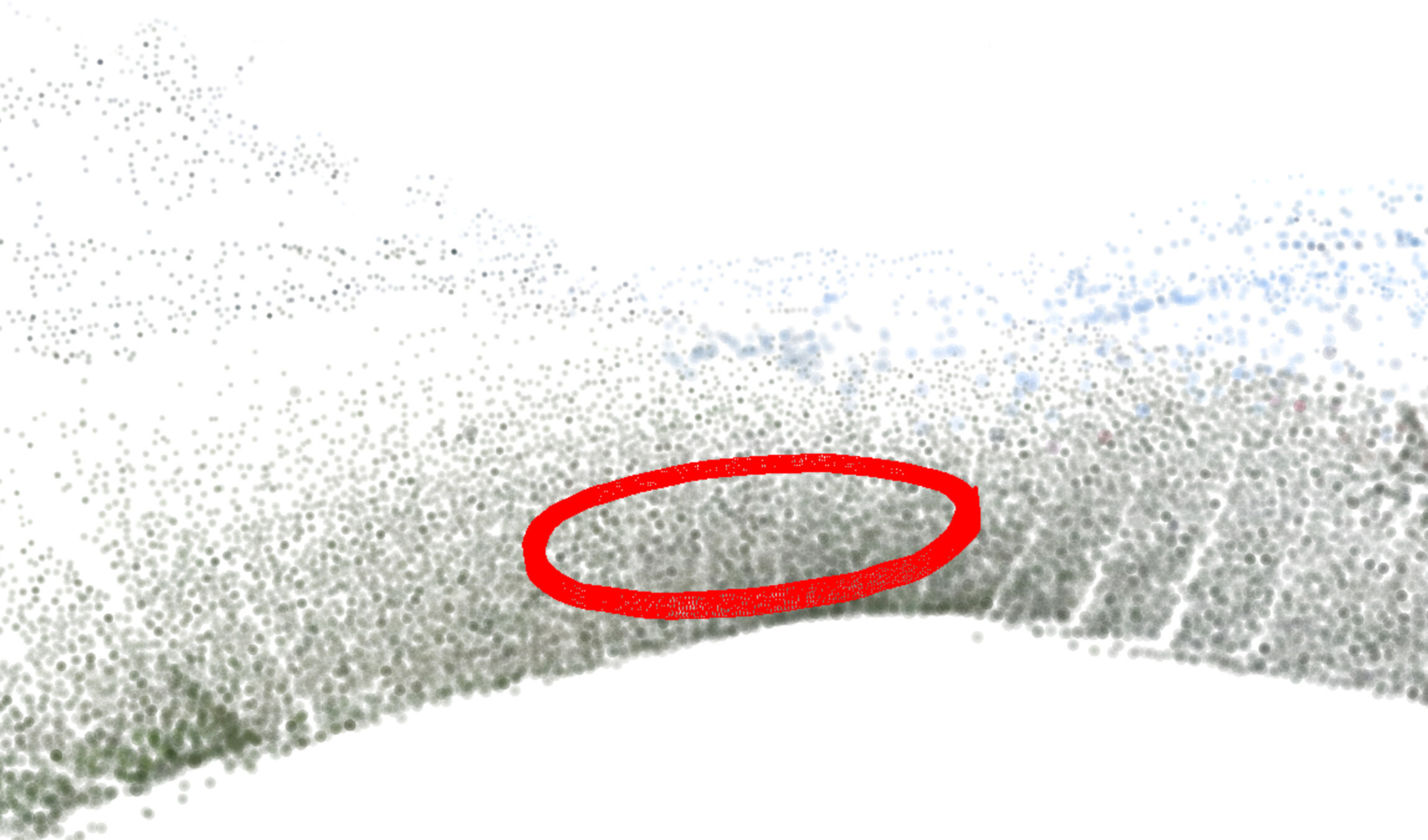}
\includegraphics[width=0.495\linewidth,keepaspectratio]{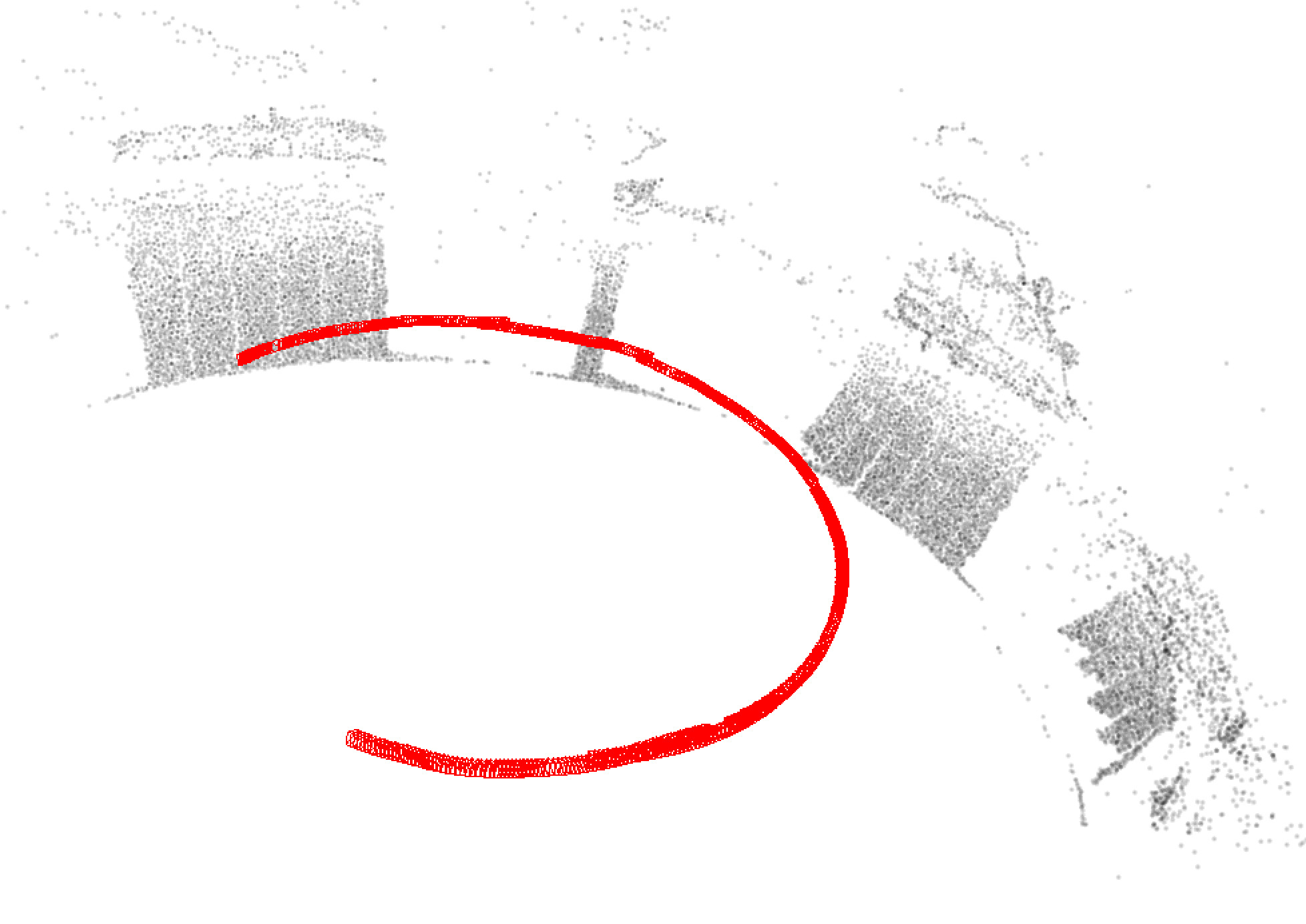}
\caption{\label{fig:sfm_calibration} Our method (left) performs accurate calibration online and reconstructs the Hawaii scene correctly. Using bad calibration (right) can lead to inaccurate pose estimates with broken structure despite an explicit loop closure detection.}
\vspace{-0.1in}
\end{figure}

For general motion, fundamental matrices may be decomposed to obtain the focal length and relative pose parameters~\cite{sturm2001focal}; however, spherical motion is exactly a critical motion for autocalibration because the optical axes of each camera intersect at a equidistant finite point (\ie, the sphere's center). As a result, the focal length cannot be determined from the fundamental matrices alone, though the fundamental matrices remain valid~\cite{sturm2001focal}. Since the fundamental matrices cannot provide camera calibration, we must resort to another method to obtain the camera focal lengths.

For our target application scenario (outward-facing handheld video), it is often the case that part of the scene is very distant. For these parts of the scene, we may approximate the relative motion between keyframes as \textit{pure rotation} instead of motion on a sphere. This is a reasonable approximation when the scene is distant because the depth of 3D point is essentially infinite and thus the translation component of the pose does not affect reprojection of the point. Further, pure rotation models have been shown to be an accurate method for camera intrinsic calibration that is capable of handling small translations~\cite{saponaro2013towards}. 

To obtain camera calibration from our uncalibrated input sequence, we estimate a pure-rotation model between pairs of keyframes in addition to fundamental matrices using the minimal solver of \cite{byrod2009minimal} to estimate relative rotation, focal length, and radial distortion. We determine whether a fundamental matrix or rotation-only motion best fits the data by computing the GRIC score~\cite{gauglitz2014model, torr2002bayesian} of both motion estimations and store the model with the best GRIC score. The focal length estimates from all rotation-only keyframe pairs are collected and a single focal length for the video sequence is computed using kernel voting~\cite{li2005non}. The radial distortion for the video sequence is similarly estimated from all fundamental matrix and rotation-only keyframes with kernel voting. Once the sequence has been calibrated, we may compute essential matrices from the fundamental matrices and extract relative pose parameters~\cite{ventura2016structure}.

\subsection{Global Scene Reconstruction}
\label{sec:sfm}



Now that we have camera calibration parameters and relative poses between keyframes, we may use the rotation averaging method of \cite{chatterjee2013efficient} to estimate global orientations for the cameras from from relative rotations. The position of cameras under spherical motion is exactly determined by the camera orientation, so this effectively provides us an initialization of the full camera pose for bundle adjustment. Note that we initialize all keyframe positions according the spherical motion assumption, even if the keyframe only observed rotation-only motions.

After initial pose estimation, 3D structure is estimated using inverse depth parameterization which is more stable to the small baselines observed for outward-facing reconstructions~\cite{ventura2016structure}. We then apply a series of nonlinear optimizations to refine camera poses and inverse depth structure. In order to properly constrain the camera poses and remain robust to small baselines,  we alternate between performing bundle adjustment with a soft prior to encourage spherical motion (we call this ``spherical bundle adjustment") and standard unconstrained bundle adjustment. The prior is a simple L2 loss penalizing the difference between current camera positions and the positions projected onto a unit sphere. Note that this is different from the spherical bundle adjustment of ~\cite{ventura2016structure}, where a hard constraint is used to enforce poses to adhere to spherical motion. Unlike ~\cite{ventura2016structure}, our bundle adjustment scheme allows for deviations from spherical motion but still prevents the camera poses from collapsing into a single point.

We use the following strategies for optimization:

\textbf{A:} Spherical bundle adjustment, no bad point filtering.

\textbf{B:} Spherical bundle adjustment,  bad point filtering.

\textbf{C:} Unconstrained bundle adjustment, no intrinsics optimization, bad point filtering

\textbf{D:} Unconstrained bundle adjustment, intrinsics optimization, bad point filtering.

We use the following sequence to refine camera poses and 3D structure: ABABCD. The initial camera poses may be noisy, so we first run bundle adjustment while constraining the motion to a sphere to allow for a well-constrained optimization when potentially few 3D points are triangulated (A). This substantially improves the initial poses, and allows for more points to be triangulated (B) and a better conditioned optimization. We then run unconstrained bundle adjustment following by a final round of camera intrinsics optimization. Successive rounds of triangulation, point filtering, and optimization allow for a wide convergence radius when the initial poses are inaccurate while simultaneously allowing for deviations from a a purely spherical motion model.

\section{Experiments}
We conducted several experiments using synthetic and real data to measure the performance of our algorithms, and  compare them to state of the art solutions for general motion. For the 4-point algorithm for estimating fundamental matrices, we compare against the 8-point algorithms~\cite{hartley2003multiple} provided in the TheiaSfM library~\cite{theia-manual}. For the 5 and 6-point algorithms which estimate fundamental matrices and radial distortions, we compare against Fitzgibbon's 9-point method~\cite{fitzgibbon2001simultaneous}.

In order to generate synthetic data for our tests, we generate random spherical motion such that the relative rotation magnitude is between $\left[ 0, 10\degree \right]$. The camera positions are computed such that both cameras lie on the unit sphere. We generate 1000 3D points  randomly with a depth between $[6,10]$ with respect to the first camera. Each point is projected into both cameras using a focal length of 1200. 

\subsection{Timing}
\begin{table}[t]
\begin{center}
 \caption{\label{table:timings} We measured the mean runtime of our solvers over 10,000 trials. Our 4-point algorithm is of comparable efficiency to the 8-point algorithm~\cite{hartley2003multiple}, while our 6-point algorithm is 5$\times$ faster than the 9-point algorithm~\cite{fitzgibbon2001simultaneous} and 2 orders of magnitude faster than our 5-point algorithm.}
\begin{tabular}{| c  | c |}
\hline
Method & Time ($\mu$s) \\ \hline
4-point Fundamental  & 4.34 \\
5-point Fundamental + $\lambda$ & 1188.46 \\
6-point Fundamental + $\lambda$  & 9.92 \\ \hline
3-point Spherical Essential~\cite{ventura2016structure} & 6.9 \\
8-point Fundamental~\cite{hartley2003multiple} & 4.11 \\
9-point Fundamental + $\lambda$~\cite{fitzgibbon2001simultaneous} & 51.91 \\ \hline
\end{tabular}
\end{center}
\vspace{-0.1in}
\end{table}

We calculated the average computation time for our solvers over 10 000 randomly
generated problems. The testing was performed on a 3.4 GHz Intel Core
i7 with code written in C++. The mean runtimes of our algorithms are given in Table~\ref{table:timings}. Our 4-point and 6-point solvers are extremely efficient, with runtimes comparable to or better than the comparable algorithms for general motion. The proposed 5-point algorithm is 2 orders of magnitude slower, indicating that it is not suitable for real-time use.

\subsection{Numeric Stability}
\begin{figure}[t]
\centering
\includegraphics[width=0.495\linewidth,keepaspectratio]{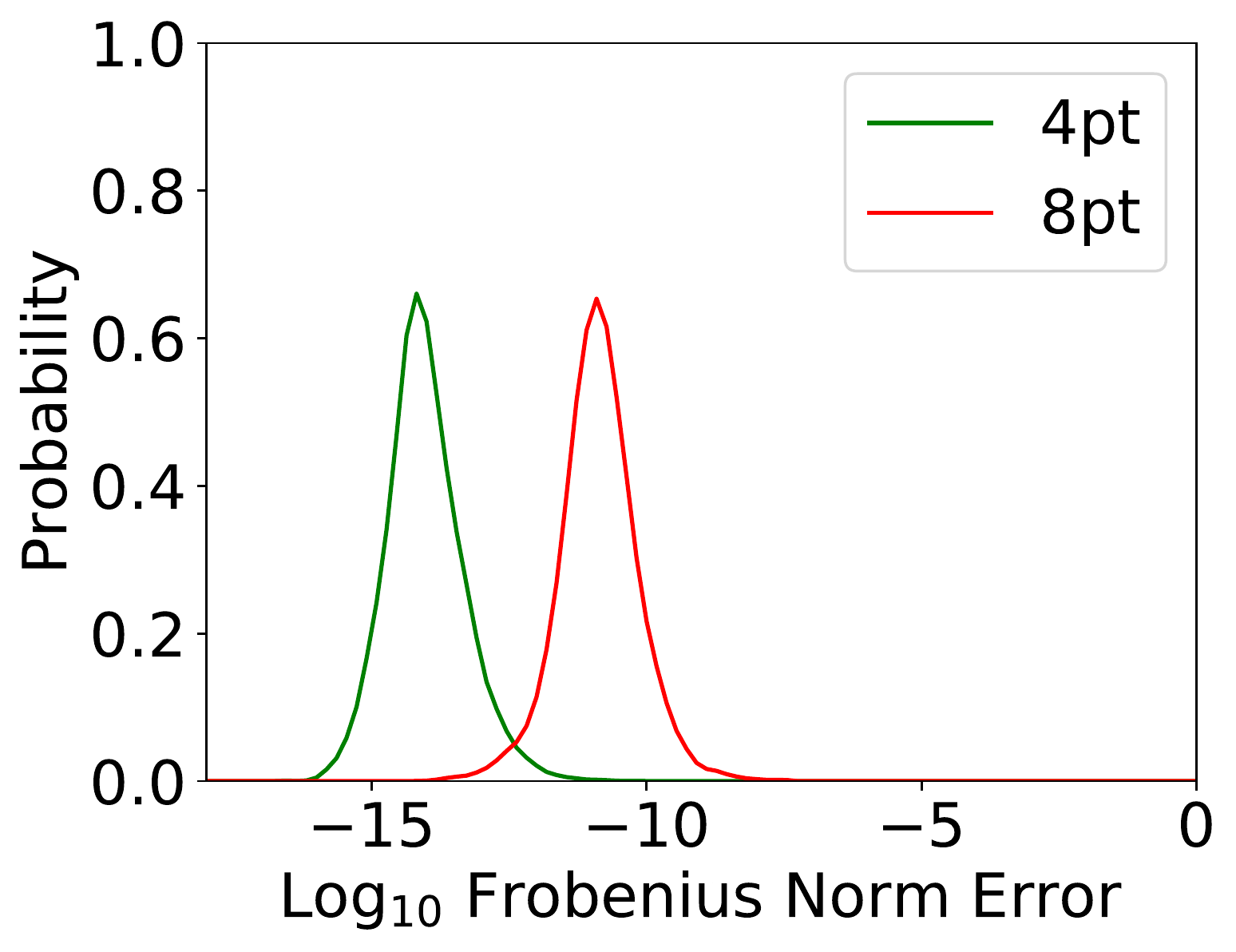} \\
\includegraphics[width=0.495\linewidth,keepaspectratio]{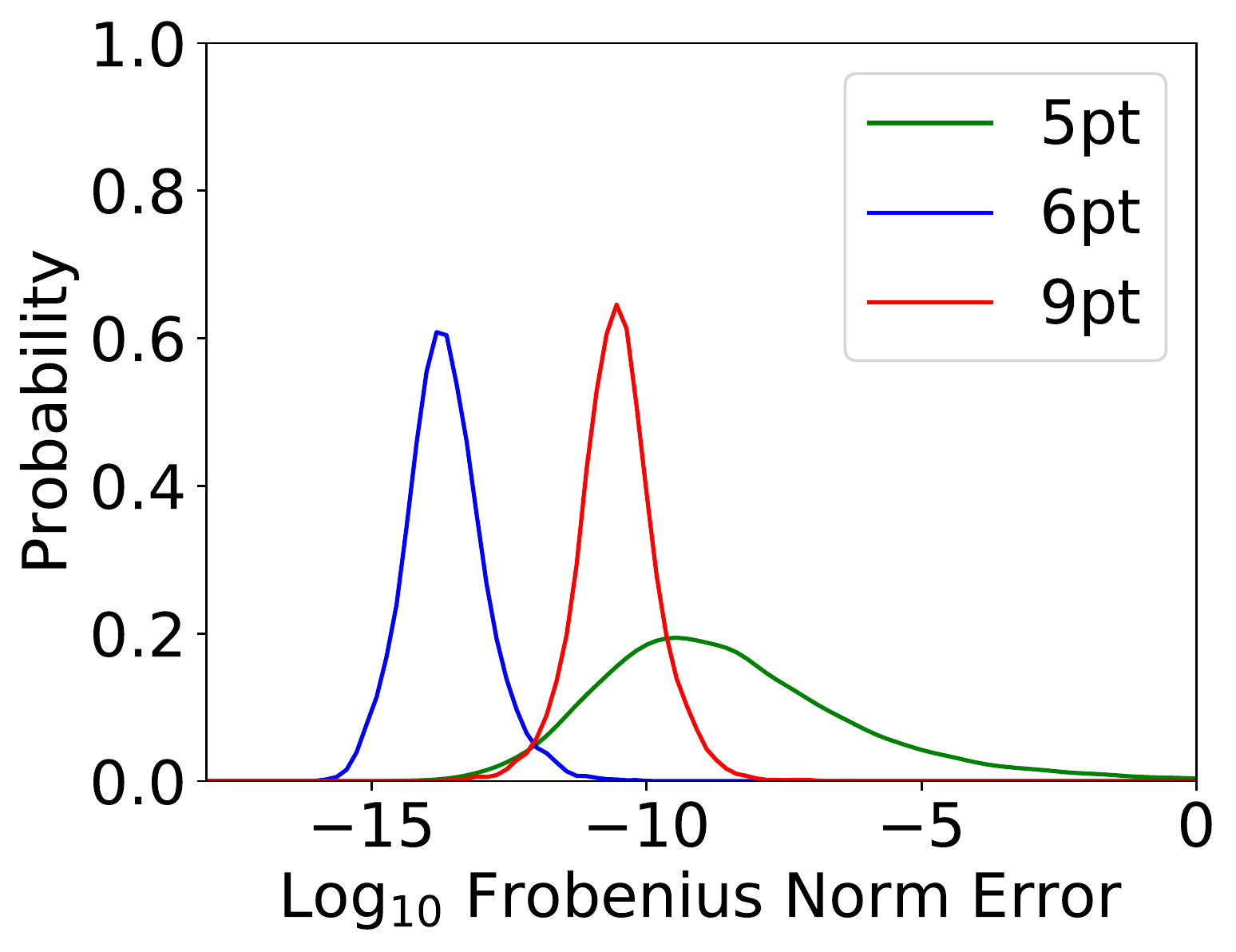}
\includegraphics[width=0.495\linewidth,keepaspectratio]{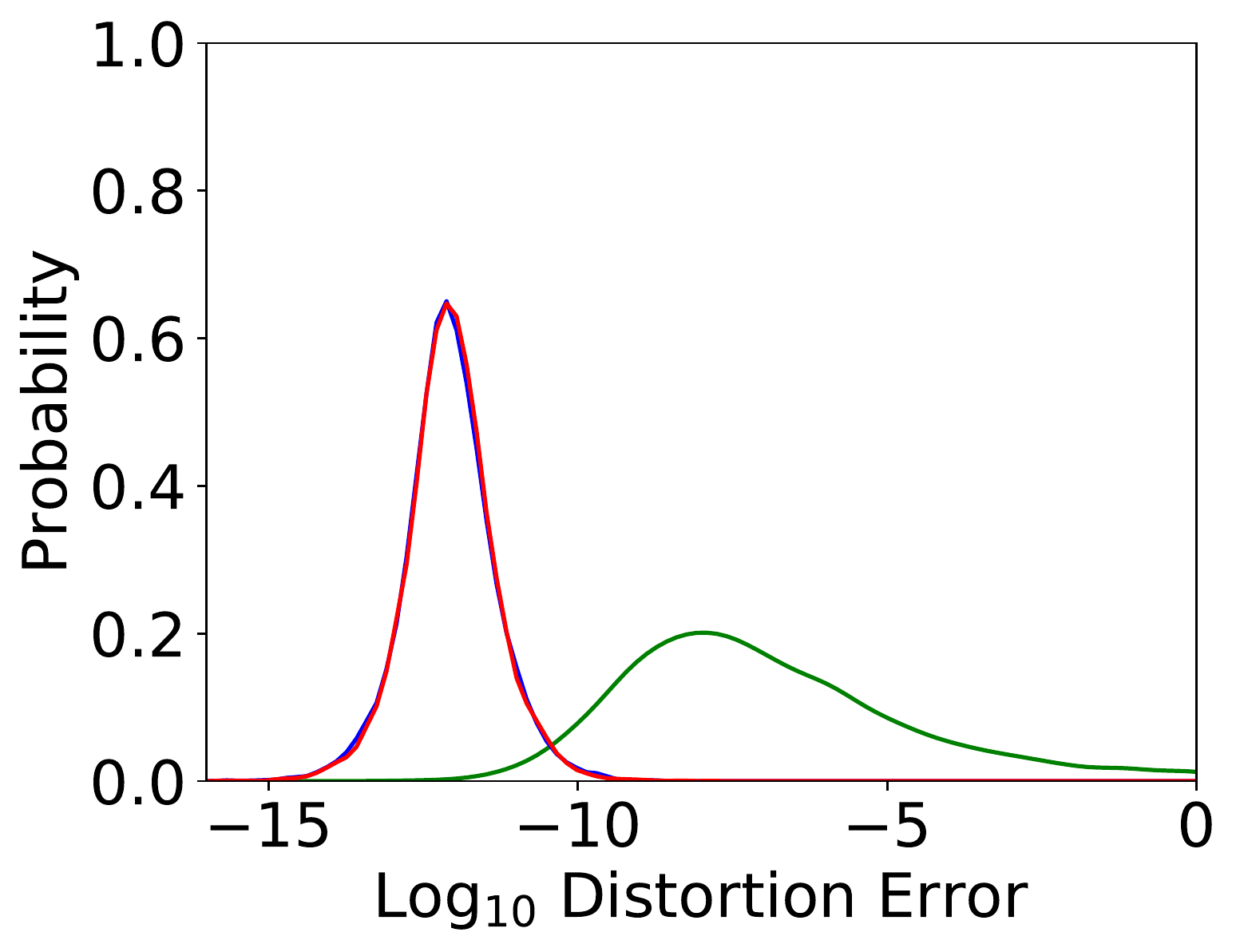}
\caption{\label{fig:num_stability_with_distortion} We measure the numeric stability of our spherical motion solvers compared to the general motion counterparts. \textbf{Top:} Our 4-point algorithm is 2-4 orders of magnitude more accurate than the 8-point algorithm~\cite{hartley2003multiple}. \textbf{Bottom:} Our 6-point algorithm exhibits comparable or better numeric stability than the 9-point algorithm~\cite{fitzgibbon2001simultaneous} while our 5-point algorithm is the least accurate.}
\vspace{-0.1in}
\end{figure}

We tested the numerical accuracy of the solvers with ideal, zero-noise observations. We generated 10000 random problems using the configuration described above, and measure the Frobenius norm of the error between the estimated and true fundamental matrix. The normalized distortion error is measured for our 5- and 6-point algorithms, along with the 9-point algorithm~\cite{fitzgibbon2001simultaneous} as: $|\lambda -\lambda_gt|/|\lambda_gt|$. The results are plotted in Figure~\ref{fig:num_stability_with_distortion}. The 4-point and 6-point solvers demonstrate excellent numerical stability, with 98\% of the fundamental matrix errors below $10^{-12}$. The 5-point algorithm is notably less stable in computing both the fundamental matrix and radial distortion, with the 6-point solver being 2-4 orders of magnitude more accurate.

\subsection{Robustness to Image Noise}
\begin{figure}[t]
\centering
\includegraphics[width=0.495\linewidth,keepaspectratio]{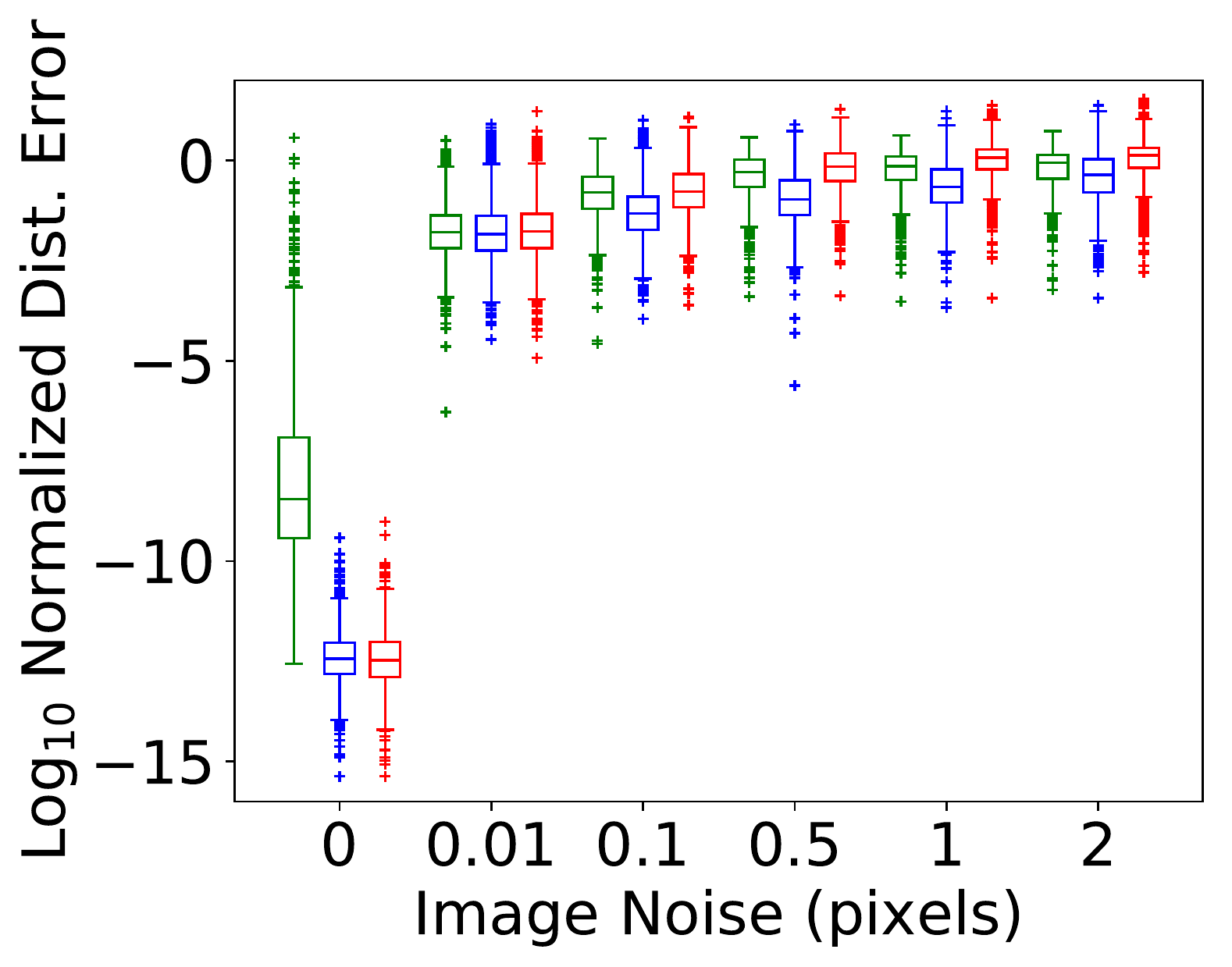}
\includegraphics[width=0.495\linewidth,keepaspectratio]{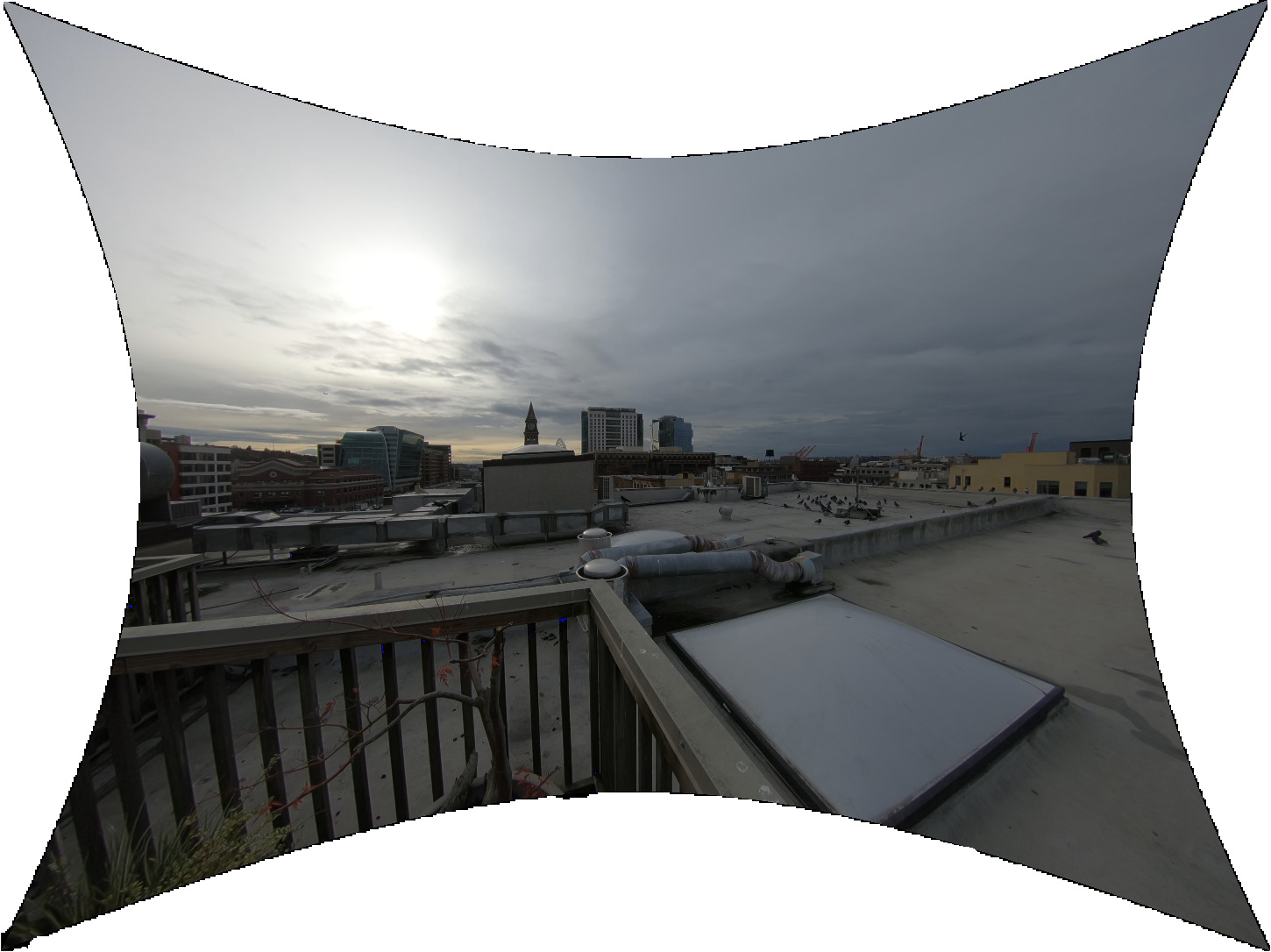}
\caption{\label{fig:image_noise}\textbf{Left:} Our 5-point algorithm (green) and 6-point algorithm (blue) perform comparably to or better than the 9-point algorithm~\cite{fitzgibbon2001simultaneous} (red) under image increasing noise, though the 5-point algorithm has the worst performance for zero-noise. \textbf{Right:} Our system successfully undistorts images in the \textit{Rooftop} sequence captured with a wide angle GoPro camera.}
\vspace{-0.1in}
\end{figure}

To measure the accuracy and robustness of our methods,
we performed an experiment on synthetic data with Gaussian
pixel noise ranging from 0 to 2 pixels. We performed
1,000 trials at each level of pixel noise. For each trial, we generate using the random setup described above. When multiple solutions are returned by the minimal solvers, we use a single additional point to select the best solution among them.

The results are plotted in Figure~\ref{fig:image_noise}.  Our 4-point algorithm for computing fundamental matrices demonstrates comparable robustness to the standard 8-point algorithm~\cite{hartley2003multiple}. The accuracy of determining radial distortion parameters is known to be sensitive to image noise~\cite{kukelova2013real}, though our 6-point algorithm is able to compute the distortion with slightly better accuracy than both our 5-point and Fitzgibbon's 9-point algorithms. 

\subsection{SfM Pipeline}
\begin{figure}[t]
\begin{center}
\includegraphics[width=0.48\linewidth,keepaspectratio, valign=t]{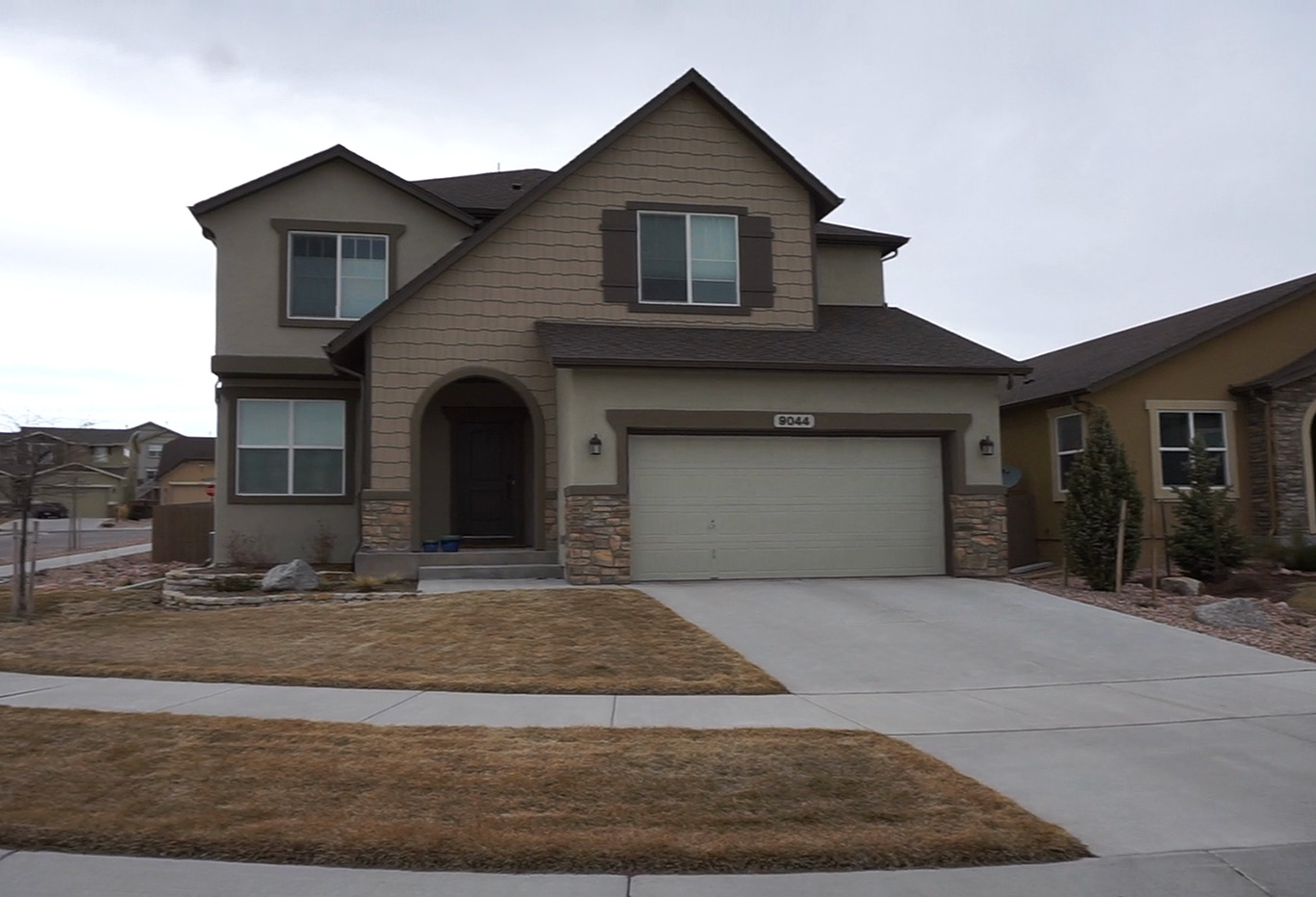} \hspace{0.1cm}
\includegraphics[width=0.48\linewidth,keepaspectratio, valign=t]{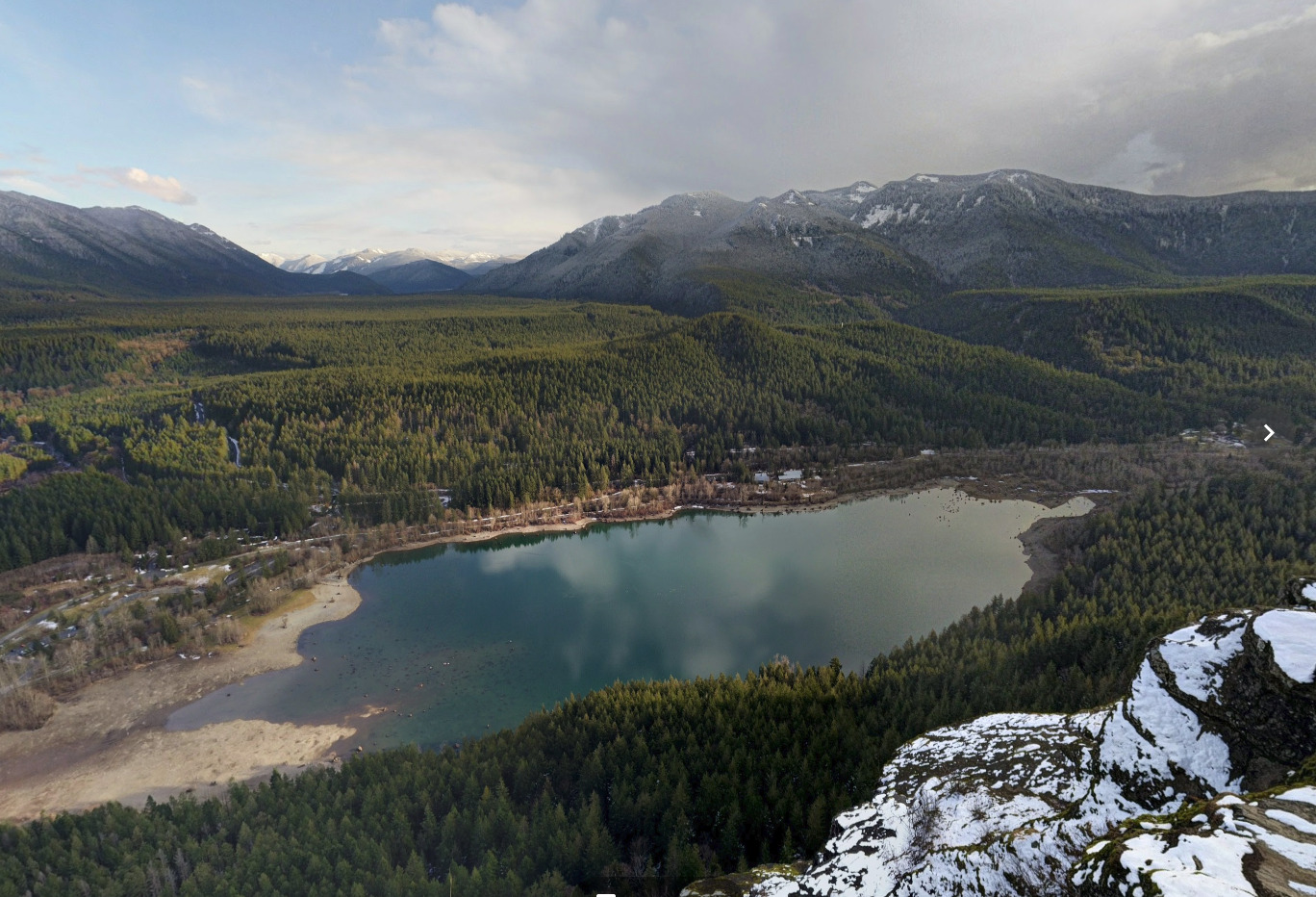} \\
\includegraphics[width=0.48\linewidth,keepaspectratio, valign=t]{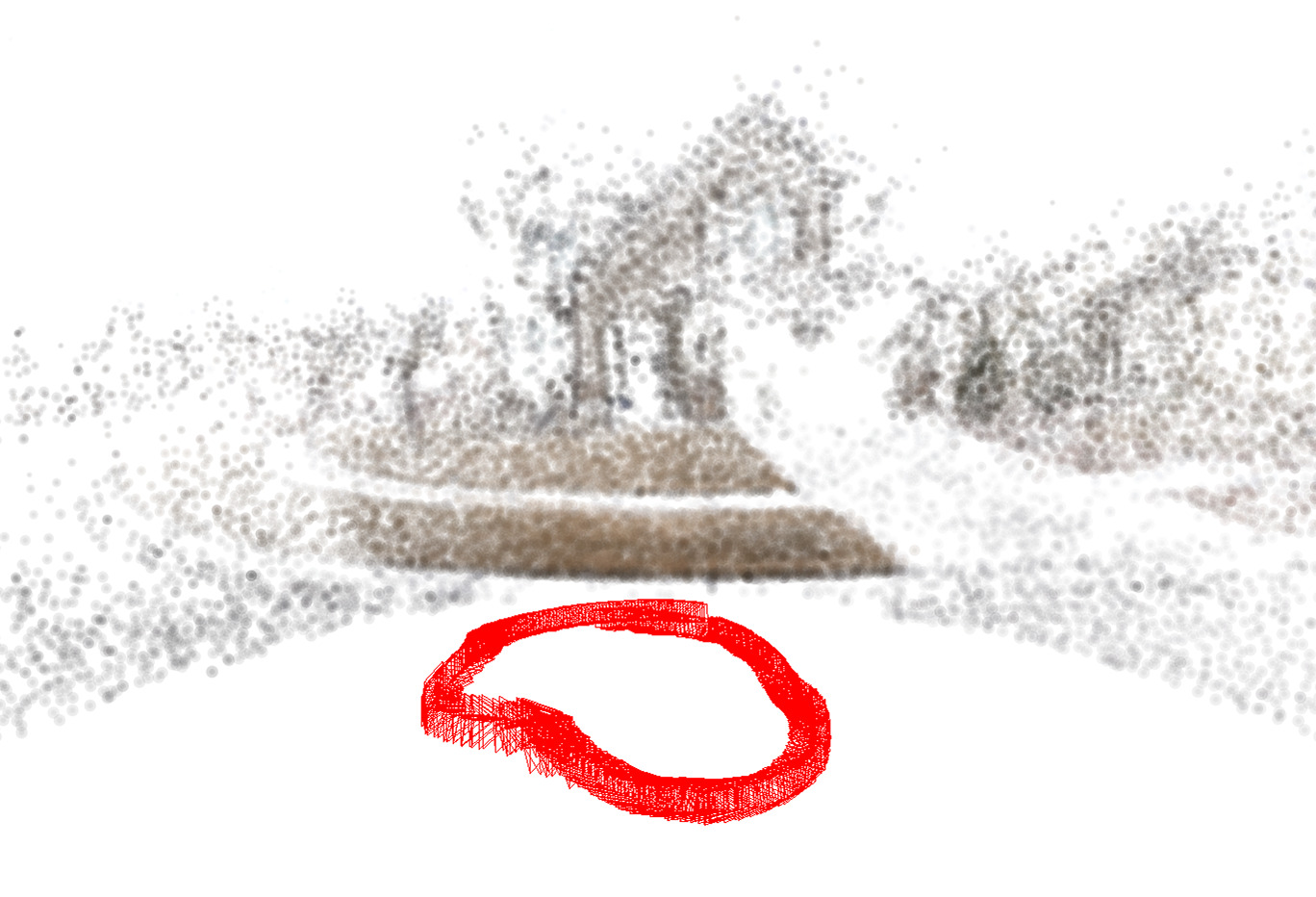} \hspace{0.1cm}
\includegraphics[width=0.48\linewidth,keepaspectratio, valign=t]{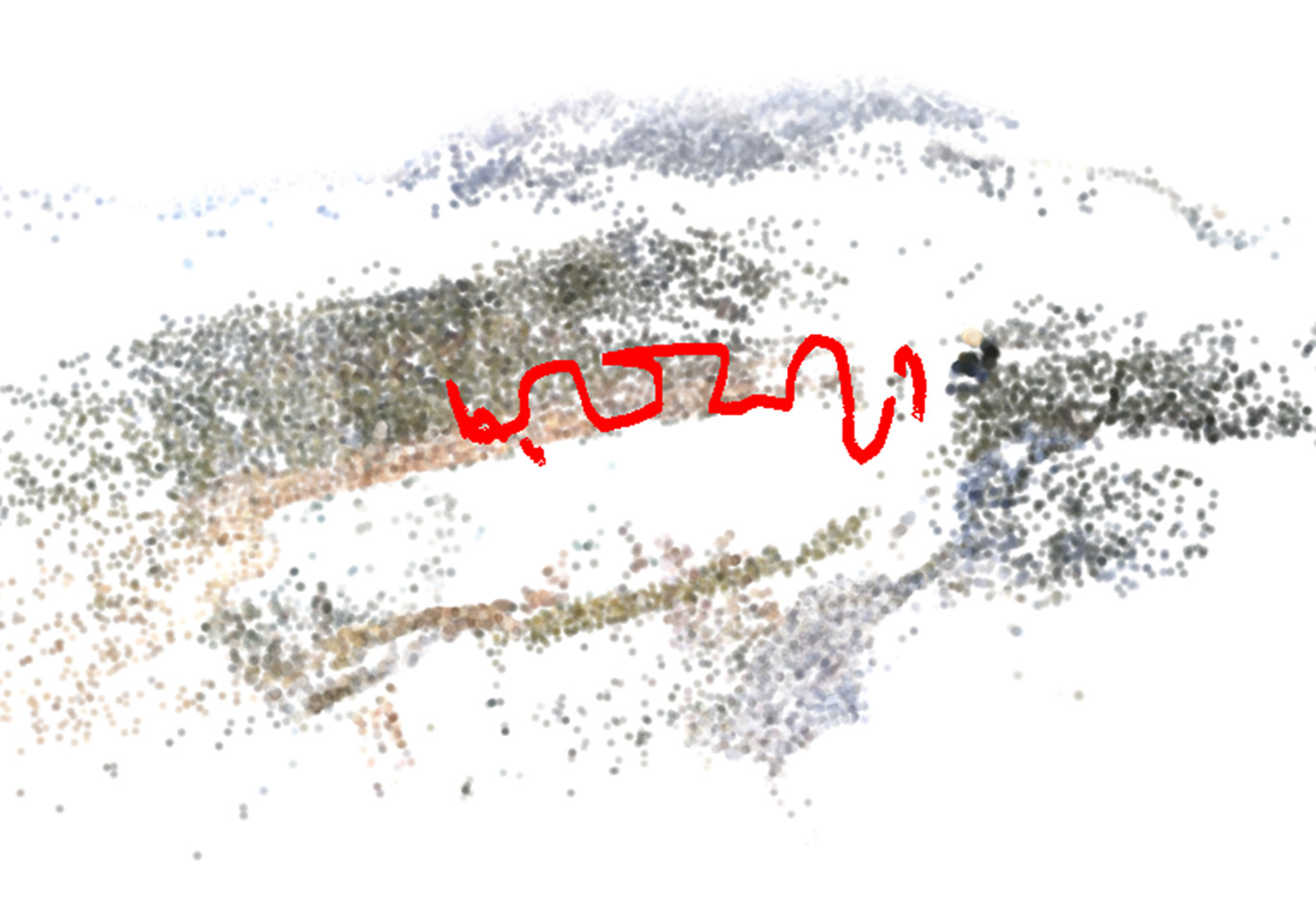}
\caption{\label{fig:sfm_results}\textbf{Left:} Our method successfully reconstructs the \textit{Street} dataset~\cite{ventura2016structure} without prior calibration. \textbf{Right:} Despite only observing distant features, our method is able to reconstruct camera poses from the \textit{Rattlesnake} dataset. Reconstructed camera frustums are shown in red.}
\end{center}
\vspace{-0.2in}
\end{figure}

\begin{table}[t]
\begin{center}
 \caption{\label{table:sfm_timings} Our SfM system is extremely efficient, and is able to reconstruct all frames in a video sequence at 10-20Hz.}
\begin{tabular}{| c | c | c | c |}
\hline
Dataset & \# Frames & Recon. Time (s) \\ \hline
Street~\cite{ventura2016structure} & 465 & 42\\
Hawaii & 1481 & 86 \\
Rattlesnake & 1864 & 102 \\
GoPro & 1343 & 112 \\
\hline
\end{tabular}
\end{center}
\vspace{-0.2in}
\end{table}

We tested the SfM pipeline described in Section~\ref{sec:sfm} on several video sequences to demonstrate the efficiency and robustness of our system, using the proposed 6-point algoritihm to estimate fundamental matrices and radial distortion parameters. The datasets were captured using a Google Pixel phone while holding the phone at arm's length facing outward. These datasets include a variety of outdoor scenes with near and distant objects. The video lengths range from 1 minute to three minutes. The \textit{Hawaii} dataset captures a scene at the beach with mountains nearby with a full 360$\degree$ loop. \textit{Rattlesnake} was captured at the cliff of a mountain looking down into a valley, with the camera making a sinusoidal motion at arm's length. The \textit{Rooftop} dataset captures a 360$\degree$ loop with a GoPro Hero 5 camera that has significant radial distortion. We additionally test the outward-facing \textit{Street} dataset of~\cite{ventura2016structure}. For all datasets we only provide the video as input to our system without any calibration. 

The limited field of view of the camera and the small baseline between views makes these videos challenging to reconstruct. Motion blur and rolling shutter effects introduce additional challenges for tracking and keypoint localization. Despite these challenges, our novel spherical motion solvers provide an excellent initialization of camera intrinsic parameters and poses for inverse-depth bundle adjustment. The spherical motion prior (\cf Section~\ref{sec:sfm}) enforced during nonlinear optimization allows the camera poses to be well-constrained in the absence of well-triangulated 3D structure. As a result, our system is able to accurately reconstruct all frames in the video sequences including transitions between near and distant scene points. The speed of our relative pose solvers allows our method to be extremely efficient and we are able to reconstruct video frames at a rate of 10-20Hz (\cf Table~\ref{table:sfm_timings}). Figures ~\ref{fig:sfm_results} show results of our SfM reconstructions.

\begin{figure}[t]
\centering
\includegraphics[width=1.0\linewidth,keepaspectratio]{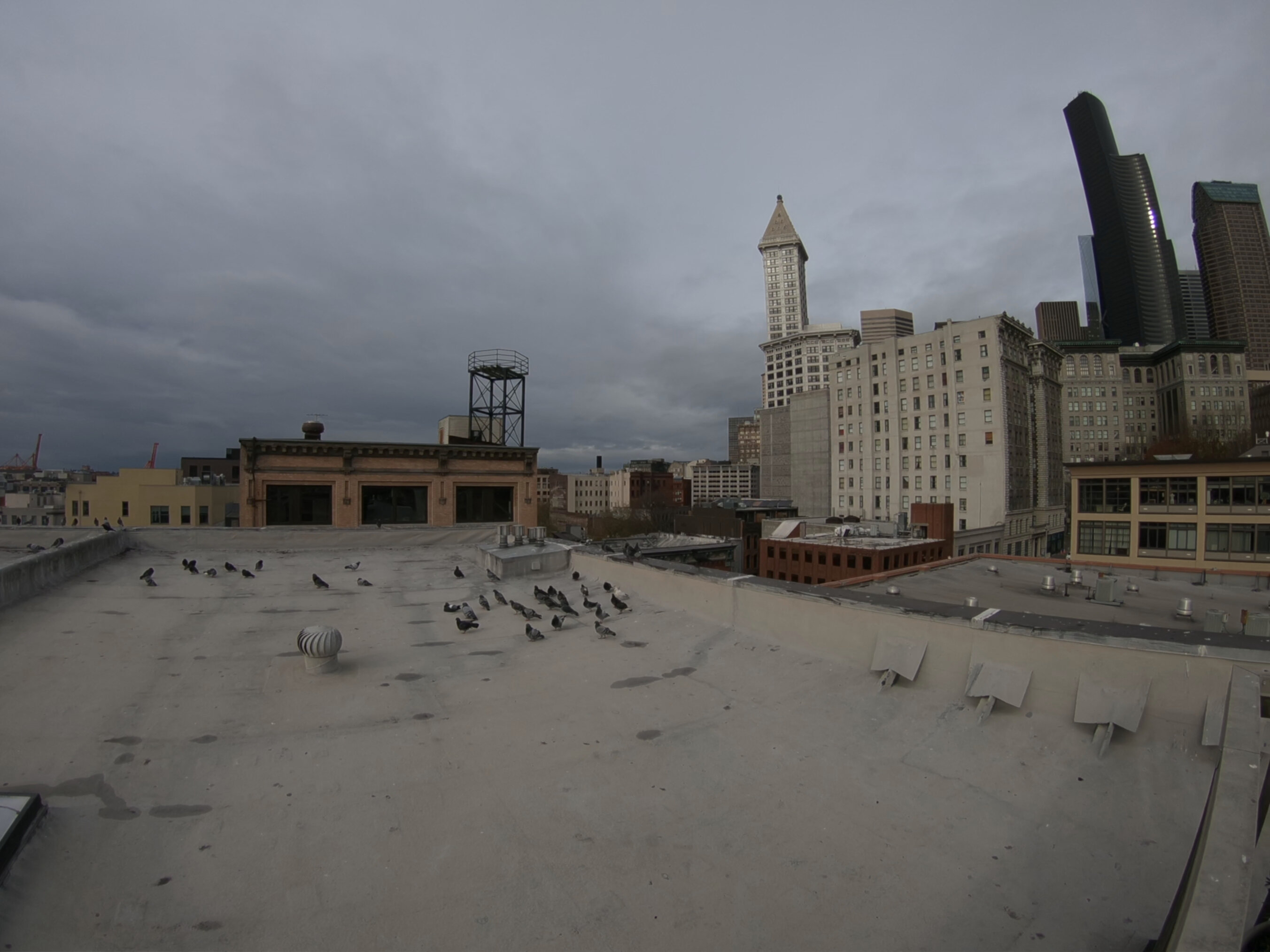}
\includegraphics[width=1.0\linewidth,keepaspectratio]{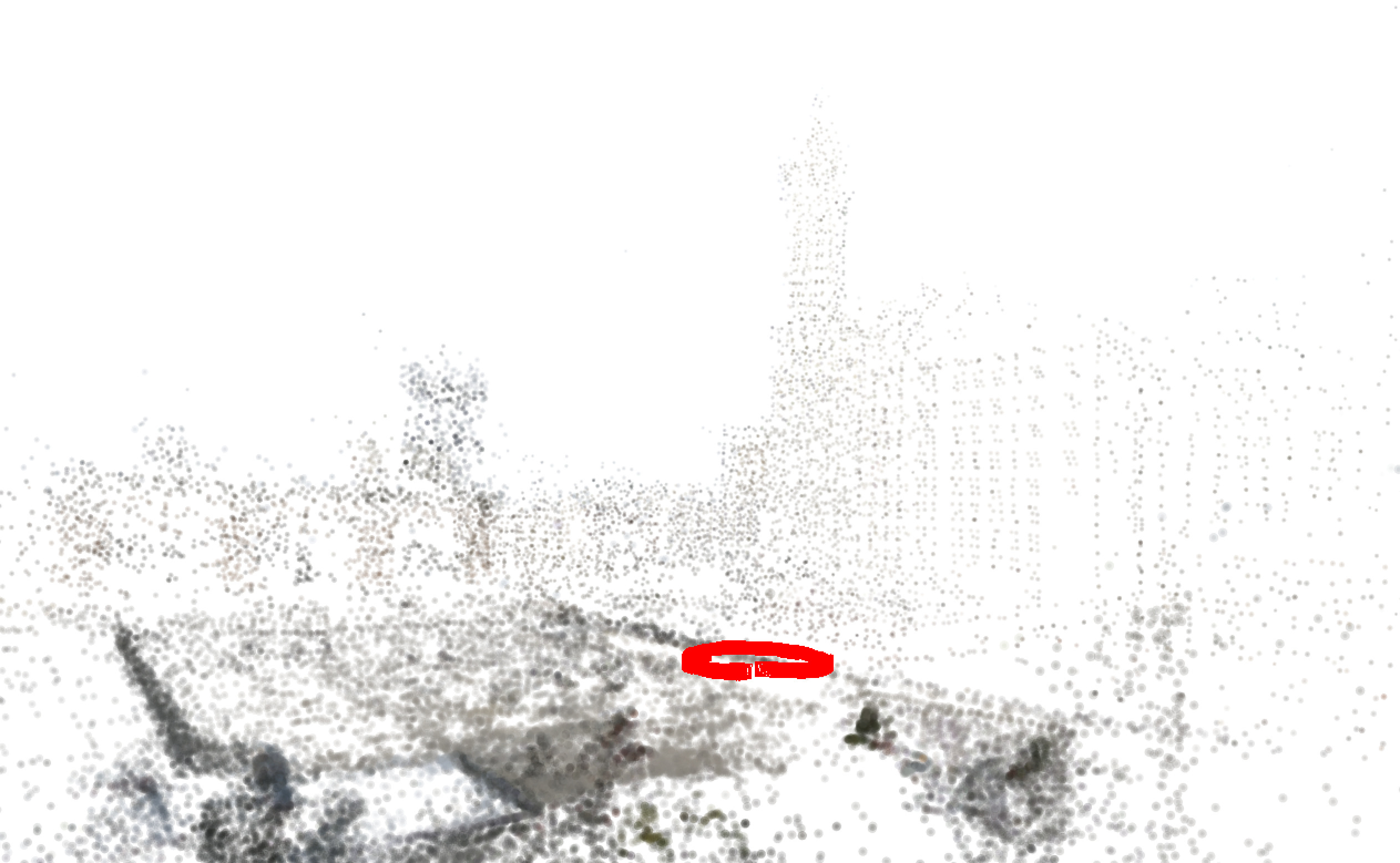}
\caption{The \textit{Rooftop} sequence is captured with a GoPro Hero5 camera which features a wide-angle lens with significant radial distortion. Given an uncalibrated video as input, our method computes accurate structure of the rooftop and the surrounding buildings.}
\vspace{-0.1in}
\end{figure}

\begin{figure}[t]
\centering
\includegraphics[width=1.0\linewidth,keepaspectratio]{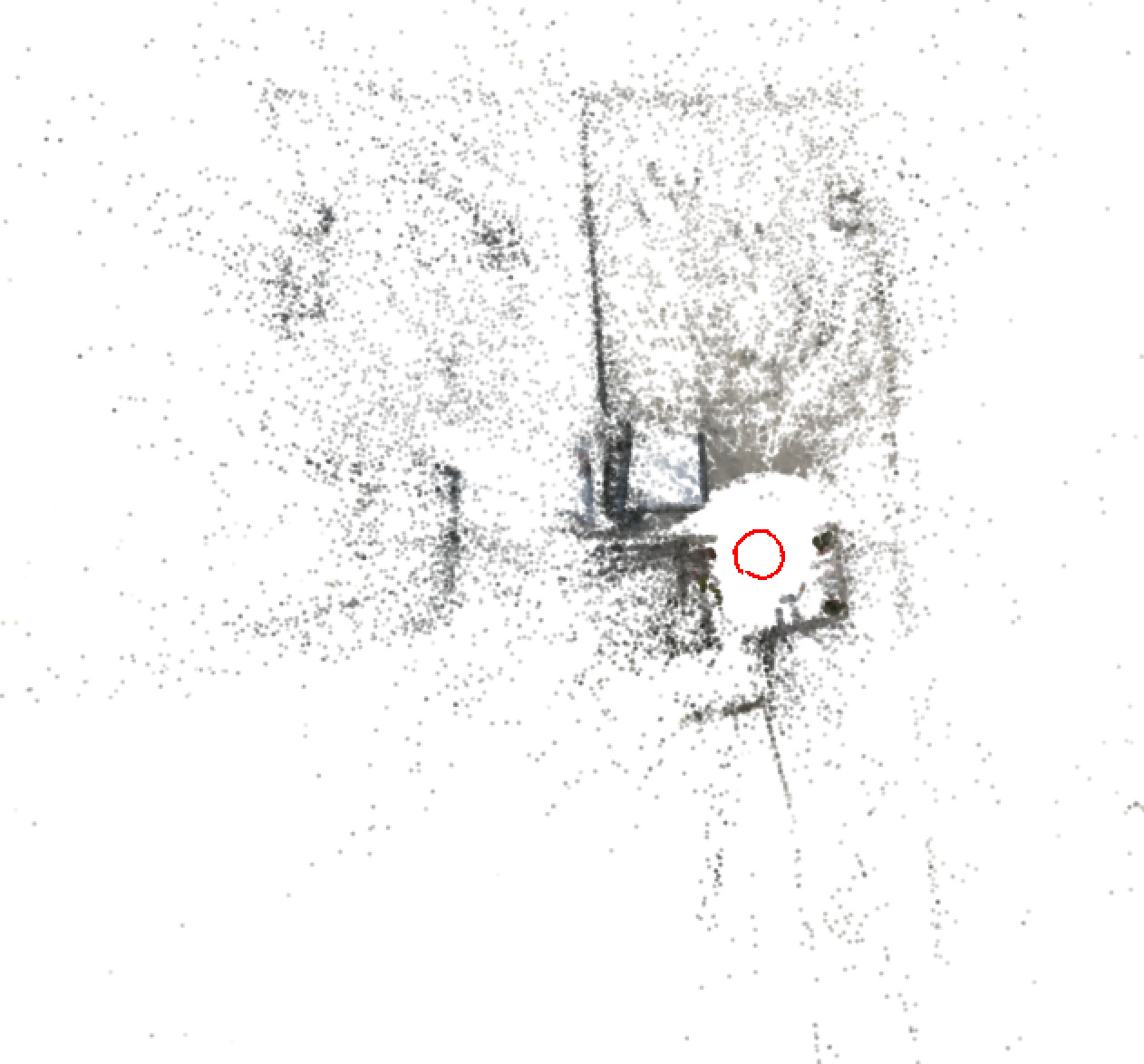}
\caption{An overhead view of the SfM reconstruction of the \textit{Rooftop} video sequence.}
\vspace{-0.1in}
\end{figure}

We attempted to reconstruct the video datasets with the open-source 3D reconstruction libraries COLMAP~\cite{schoenberger2016sfm}, Theia~\cite{theia-manual}, and ORB-SLAM2\footnote{\url{https://github.com/raulmur/ORB_SLAM2}}. ORB-SLAM2 requires calibrated input so we have provided camera intrinsic parameters obtained with OpenCV's checkerboard calibration. None of these systems were able to reconstruct any of the video sequences successfully, while our system successfully reconstructs all video sequences without any prior calibration information provided. Even when providing calibration parameters to COLMAP and Theia both systems fail to obtain reconstructions for any of the input datasets. This is a significant result, as our system allows for accurate reconstruction where traditional SfM and SLAM systems are unable to obtain even partial reconstructions. Accurate pose reconstruction enables several interesting applications such as depth-aware panorama reconstruction~\cite{hedman2017casual} or dense reconstruction with multiview stereo~\cite{furukawa2010accurate, schonberger2016pixelwise}.

\section{Multiview Stereo Reconstruction}

\begin{figure}[t]
\centering
\includegraphics[width=1.0\linewidth,keepaspectratio]{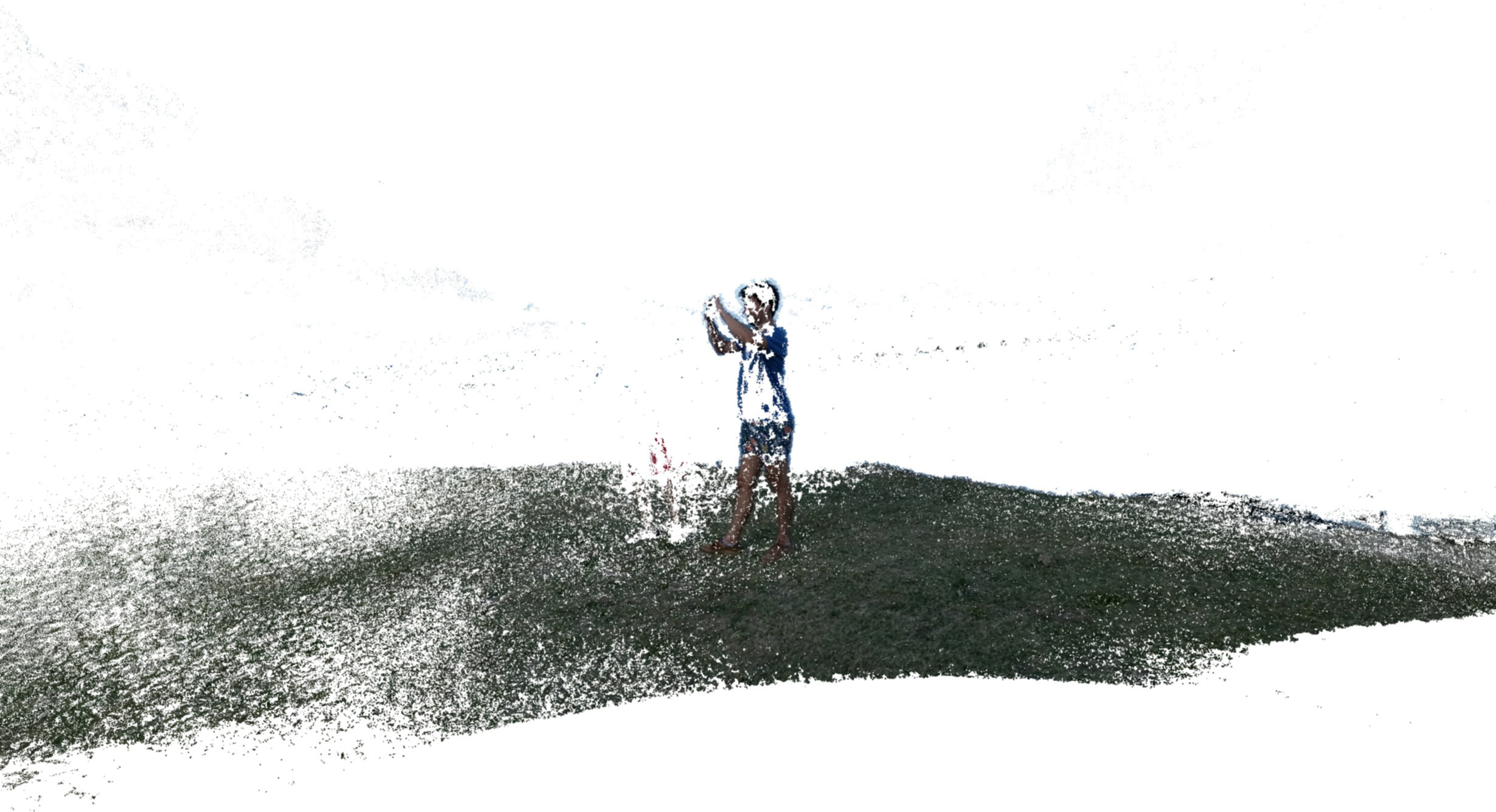}
\caption{Accurate camera poses recovered from our system enables dense reconstruction with multiview stereo~\cite{schonberger2016pixelwise}}
\vspace{-0.1in}
\end{figure}

A motivating application for our system is to enable high quality multiview stereo (MVS) scene reconstructions, by simply rotating a camera on a circle.
Panorama-style capture is one of the most common and convenient methods for capturing a scene, yet most SfM systems fail to reconstruct poses for sequences with this type of motion.  

\begin{figure}[t]
\centering
\includegraphics[width=0.475\linewidth,keepaspectratio]{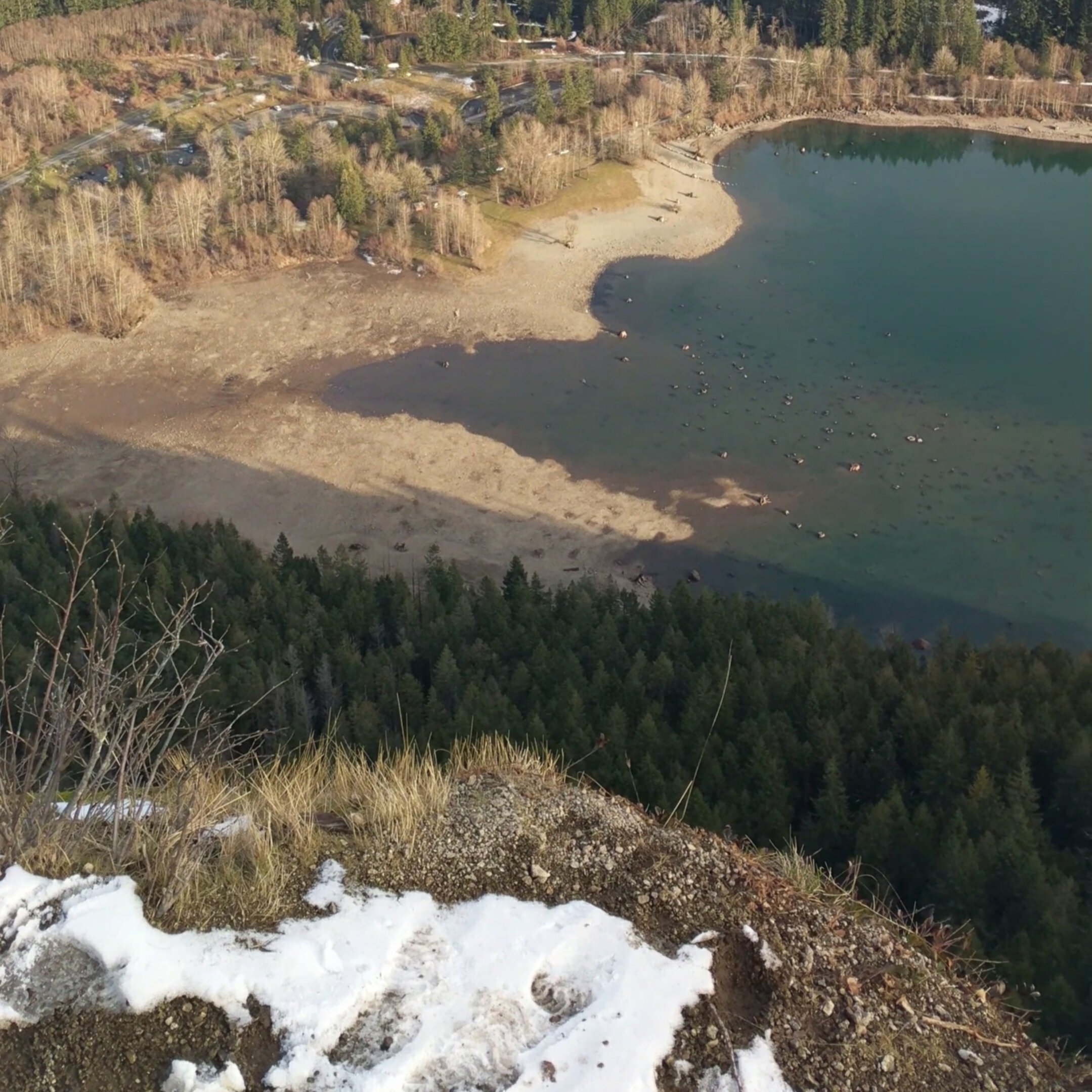}
\includegraphics[width=0.475\linewidth,keepaspectratio]{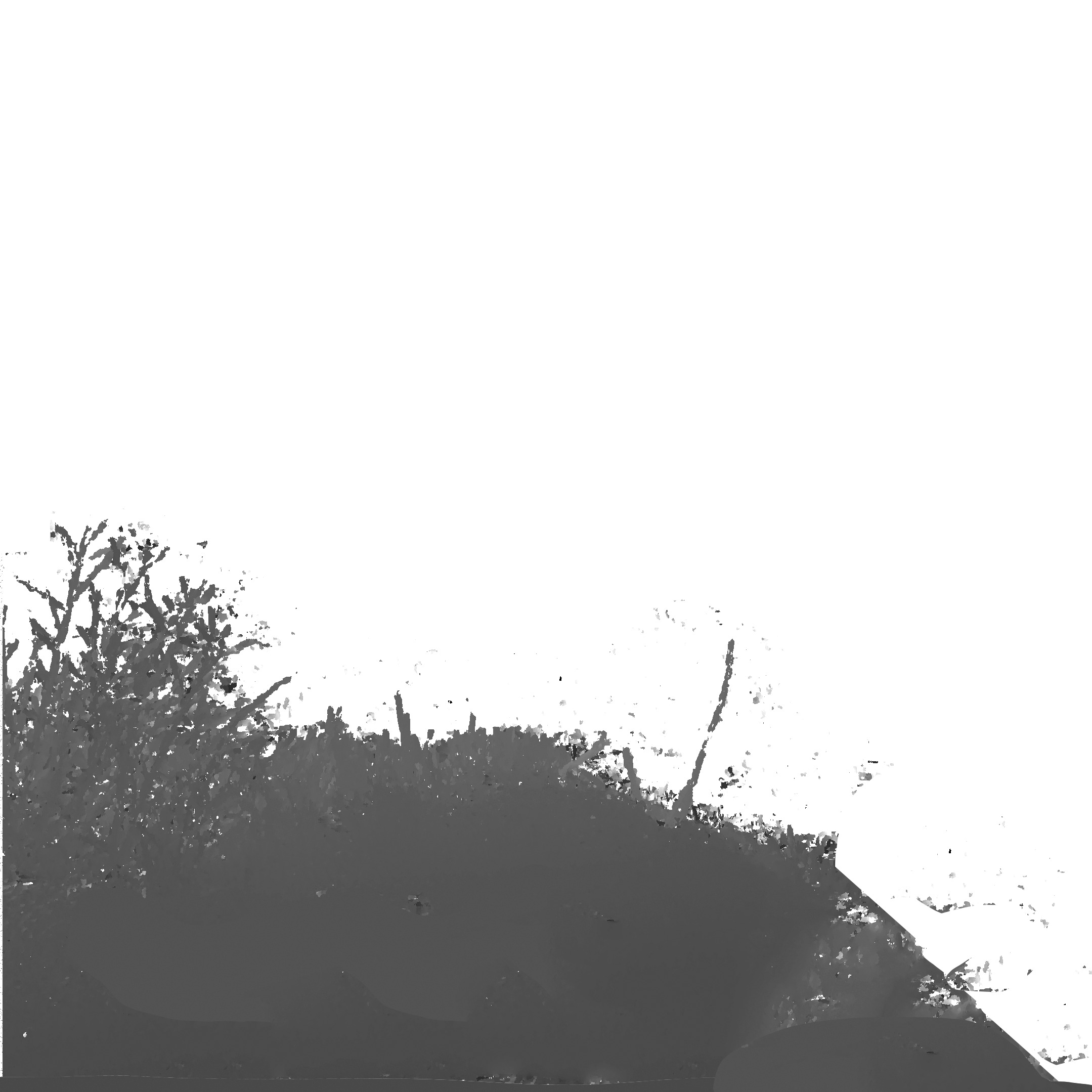}
\caption{\label{fig:rgb_depth} An input frame from the video sequence (left) and the corresponding depth map from an MVS pipeline (right). Accurate poses from our SfM system allow for depth reconstruction of near points in the scene, including thin structures, allowing for correct parallax \eg, when viewed in a VR headset.}
\vspace{-0.1in}
\end{figure}

We demonstrate that our method can yield high quality MVS reconstructions by using the camera poses and calibration obtained with our reconstruction method for an off-the-shelf MVS method~\cite{schonberger2016pixelwise}.  Note that MVS methods require highly accurate (sub-pixel) pose to reconstruct fine structures.
The MVS technique successfully extract depths for near parts of the scene, including thin objects (see Figure~\ref{fig:rgb_depth}).

\section{Conclusion}
In this paper, we have presented a new SfM method for reconstructing camera poses that lie approximately on a sphere. Our system is suitable for reconstructing video sequences when a user is rotating the camera from an outstretched arm. This panorama-style capture is common and convenient, yet previous uncalibrated SfM and SLAM solutions are unable to obtain accurate reconstructions due to the rotation-dominant motion and distant scene points. We introduced three novel relative pose solvers that compute fundamental matrices and camera calibration under the assumption of spherical motion. These solvers require fewer correspondences than the general-motion algorithms, leading to faster convergence in RANSAC schemes. We have integrated the solvers into a robust and efficient SfM pipeline that computes accurate camera pose and 3D structure. By using a soft prior on camera motion, camera poses are properly constrained during bundle adjustment and converge to a high quality solution in the absence of well-triangulated 3D points. This allows our system to handle deviations from purely spherical motion, though significant deviations will cause our method to fail.

The results of our SfM pipeline are suitable for multiview stereo reconstruction.
The ability to recover accurate camera poses even when the cameras only view distant features is important to properly model parallax. Future work includes increasing the accuracy of the system, perhaps with photometric bundle adjustment or ``epipolar segment" bundle adjustment~\cite{herrera2014dt} to properly handle features with infinite depth, and designing stereo pipelines that are more robust to small baselines to obtain better MVS reconstructions.

{\small
\bibliographystyle{ieee}
\bibliography{vr_on_sphere}
}

\end{document}